\documentclass[10pt,letterpaper,twocolumn,twoside,journal]{IEEEtran}

\usepackage{amsmath,amsfonts,amssymb,mathtools}
\usepackage[utf8]{inputenc}
\usepackage[T1]{fontenc}
\usepackage{microtype}
\usepackage{textcomp}

\usepackage{array}
\usepackage{multirow}
\usepackage{booktabs}
\usepackage{tabularx}
\usepackage{colortbl}
\usepackage{xcolor}
\usepackage{adjustbox}
\usepackage{xspace}
\usepackage{enumitem}
\usepackage{graphicx}
\IfFileExists{stfloats.sty}{\usepackage{stfloats}}{}
\usepackage{url}
\usepackage{cite}
\newcommand{\corrmark}{*}
\newcommand{\corrauth}{\textsuperscript{\corrmark}}
\IfFileExists{makecell.sty}{\usepackage{makecell}}{\makeatletter\makeatother}

\usepackage[caption=false,font=footnotesize]{subfig}

\usepackage[hidelinks]{hyperref}
\hypersetup{
  pdftitle={PRISM: Per-unit Kinematic Motion Generation in a Structured Latent Manifold},
  pdfauthor={Zeyu Ling, Qing Shuai, Teng Zhang, Shiyang Li, Bo Han, and Changqing Zou},
  pdfsubject={Manuscript submitted to IEEE Transactions on Multimedia},
  pdfkeywords={human motion generation, SMPL, motion representation, flow matching, long-horizon generation}
}
\IfFileExists{cleveref.sty}{%
  \usepackage[capitalize]{cleveref}%
  \crefname{section}{Sec.}{Secs.}\Crefname{section}{Section}{Sections}%
  \crefname{table}{Tab.}{Tabs.}\Crefname{table}{Table}{Tables}%
}{%
  \newcommand{\cref}[1]{\ref{##1}}\newcommand{\Cref}[1]{\ref{##1}}%
}

\newcommand{\figref}[1]{\mbox{Fig.~\ref{#1}}}
\newcommand{\tabref}[1]{\mbox{Tab.~\ref{#1}}}
\newcommand{\secref}[1]{\mbox{Sec.~\ref{#1}}}
\newcommand{\name}{PRISM\xspace}

\begin{document}

\title{PRISM: Per-unit Kinematic Motion Generation\\in a Structured Latent Manifold}

\author{Zeyu Ling, Qing Shuai, Teng Zhang, Shiyang Li, Bo Han, and~Changqing~Zou\corrauth%
\thanks{Z.~Ling, Q.~Shuai, S.~Li, and C.~Zou are with the State Key Laboratory of CAD~\&~CG, Zhejiang University, Hangzhou, China. C.~Zou is also with Zhejiang Lab, Hangzhou, China. (e-mail: zeyuling@zju.edu.cn; s\_q@zju.edu.cn; 12521030@zju.edu.cn; changqingzou@zju.edu.cn).}%
\thanks{T.~Zhang is with the Computer Animation~\&~Perception Group, Zhejiang University, Hangzhou, China. (e-mail: zhangteng0831@gmail.com).}%
\thanks{B.~Han is with the VIVID Team, ByteDance, Beijing, China. (e-mail: borishan.815@bytedance.com).}%
\thanks{Z.~Ling, Q.~Shuai, and T.~Zhang contributed equally. \corrmark~Corresponding author: C.~Zou.}}

\IEEEaftertitletext{%
\begin{center}
\vspace{0.3em}
\includegraphics[width=\textwidth]{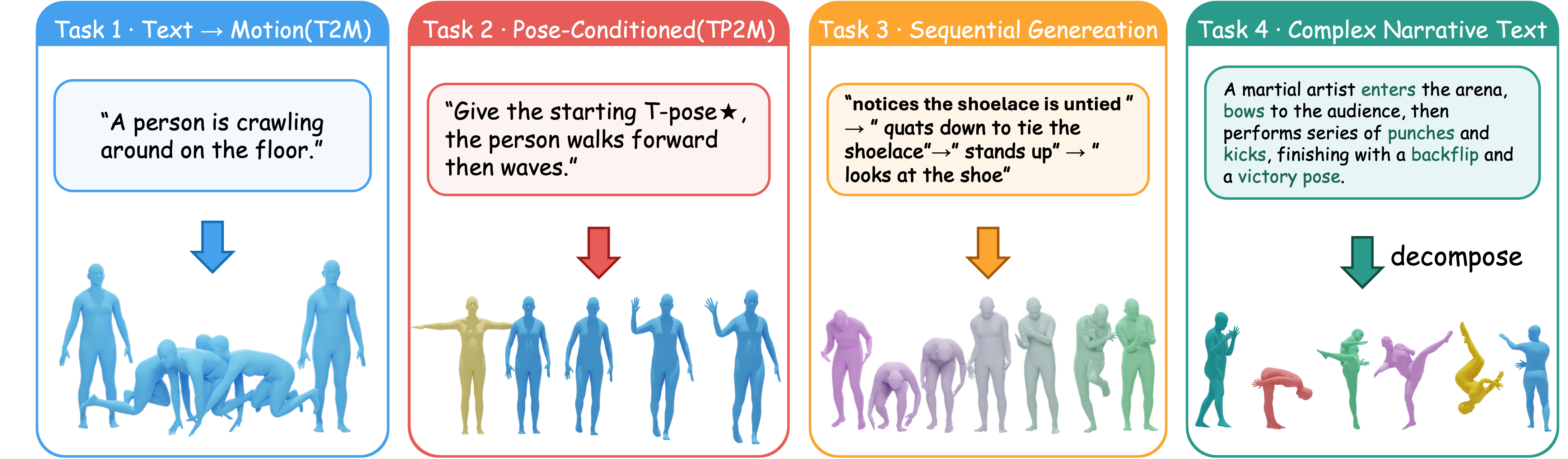}
\vspace{-0.7em}
\refstepcounter{figure}\label{fig:teaser}%
{\footnotesize Fig.~\thefigure. \name unifies text-to-motion, pose-conditioned, sequential, and narrative motion synthesis in a single model.\par}
\vspace{0.5em}
\end{center}
}

\markboth{IEEE Transactions on Multimedia}%
{Ling \MakeLowercase{\textit{et al.}}: PRISM: Per-unit Kinematic Motion Generation in a Structured Latent Manifold}

\maketitle

\begin{abstract}
Text-to-motion generation has advanced with larger corpora and stronger generators, yet many models still rely on holistic frame- or clip-level latents that entangle trajectory, orientation, and articulation. This entanglement obscures body topology and forces the generator to recover kinematic structure implicitly. We present \name, a SMPL motion generation framework that factorizes motion into continuous kinematic-unit latents. A causal Motion VAE maps motion to a time-by-kinematic-unit latent manifold, and a Kinematic-Unit Flow Transformer performs text-conditioned flow matching in this structured space. Because each latent coordinate remains tied to a physical body unit, \name can use kinematic-tree rotary position encoding and kinematic-adaptive flow scheduling. We further train the generator with per-token timesteps over clean-context/noisy-target masks, enabling frame-conditioned continuation and autoregressive segment chaining within one model. Experiments first validate the representation: the kinematic-unit VAE achieves lower geometry, rotation, and feature errors than existing motion tokenizers, showing that the latent space preserves articulated structure rather than merely compressing frames. With a 1.4B-parameter generator trained only on publicly available academic motion--text data, \name outperforms all evaluated academic-data text-to-motion baselines and remains competitive with systems trained on much larger non-public motion corpora. Without task-specific retraining, the same formulation also improves prefix-conditioned generation, BABEL sequential rollout, and narrative motion composition. These results indicate that kinematic-unit latent factorization provides an effective generation substrate for controllable SMPL motion synthesis. Code will be released at \url{https://github.com/ZeyuLing/PRISM}.
\end{abstract}


\begin{figure*}[t]
\centering
\includegraphics[width=\textwidth,keepaspectratio]{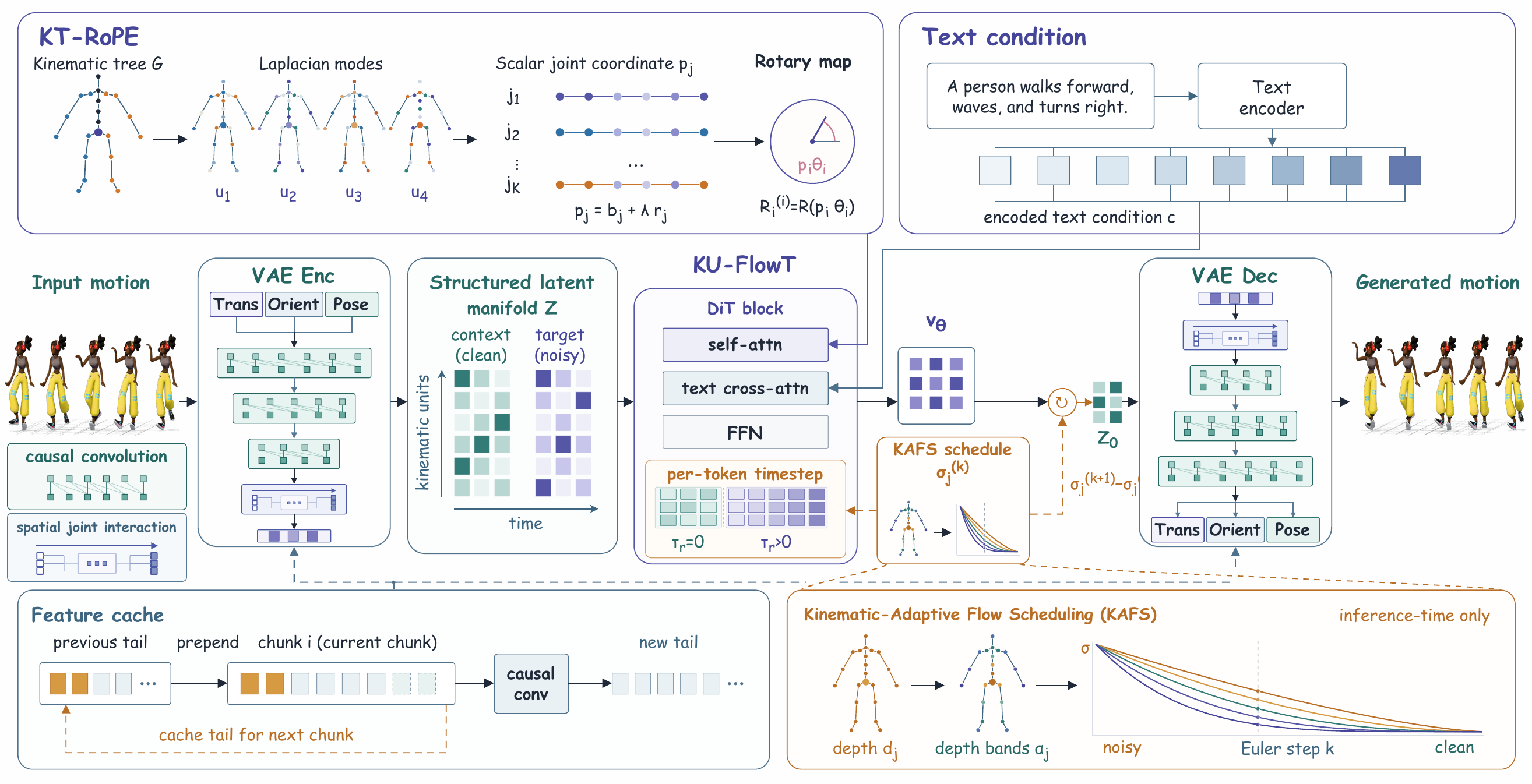}
\caption{Overview of the \name framework.}
\label{fig:pipeline}
\end{figure*}

\begin{table*}[t]
\caption{Text-to-motion on HumanML3D with the TMR evaluator.}
\label{tab:eval_t2m}
\centering
\footnotesize
\setlength{\tabcolsep}{4pt}
\resizebox{\textwidth}{!}{%
\begin{tabular}{@{}l cccccc ccc@{}}
\toprule
\multirow{2}{*}{Method}
  & \multicolumn{6}{c}{HumanML3D~\cite{t2m}}
  & \multicolumn{3}{c}{Physical Plausibility} \\
\cmidrule(lr){2-7} \cmidrule(lr){8-10}
  & R-P~T1$\uparrow$ & R-P~T2$\uparrow$ & R-P~T3$\uparrow$ & FID$\downarrow$ & MM-D$\downarrow$ & Div.$\rightarrow$
  & Slide$\downarrow$ & Float$\downarrow$ & Jitter$\downarrow$ \\
\midrule
Real (SMPL)
  & 0.770 & 0.903 & 0.944 & 0.0 & 14.88 & 27.77
  & 2.00 & 2.6 & 5.5 \\
\midrule
\multicolumn{10}{@{}l}{\emph{HumanML3D-263-based methods}} \\
Real (HML3D$\rightarrow$SMPL)
  & 0.717 & 0.872 & 0.922 & 67.4 & 16.79 & 26.63
  & 1.67 & 2.3 & 5.3 \\
MotionGPT~\cite{motiongpt}
  & 0.494 & 0.635 & 0.694 & 23.7 & 19.68 & 25.54
  & 3.88 & 10.9 & 5.2 \\
MotionGPT3~\cite{motiongpt3}
  & 0.671 & 0.824 & 0.882 & 21.0 & 17.57 & 25.69
  & 3.81 & 9.7 & 4.8 \\
MLD~\cite{mld}
  & 0.566 & 0.733 & 0.810 & 39.7 & 19.34 & 24.90
  & 8.30 & 58.3 & \underline{3.8} \\
MoMask~\cite{momask}
  & 0.640 & 0.797 & 0.861 & 21.1 & 18.12 & 25.98
  & 3.73 & 9.5 & 5.7 \\
MoGenTS~\cite{mogents}
  & 0.499 & 0.652 & 0.735 & 20.2 & 19.54 & 25.70
  & 4.15 & 12.3 & 5.1 \\
MotionLCM~\cite{motionlcm}
  & 0.566 & 0.735 & 0.808 & 44.1 & 19.45 & 24.64
  & 4.29 & 19.1 & \textbf{3.2} \\
MDM~\cite{mdm}
  & 0.521 & 0.694 & 0.770 & 35.5 & 19.42 & 25.34
  & \underline{2.53} & 5.0 & 4.7 \\
T2M-GPT~\cite{t2mgpt}
  & 0.552 & 0.706 & 0.779 & 25.5 & 19.09 & 25.59
  & 3.76 & 11.2 & 4.9 \\
FlowMDM~\cite{flowmdm}
  & 0.474 & 0.650 & 0.731 & 36.4 & 20.00 & 25.18
  & 3.05 & 7.4 & 5.0 \\
MotionLab~\cite{motionlab}
  & 0.637 & 0.788 & 0.853 & 25.4 & 17.98 & 25.54
  & \textbf{2.42} & \underline{4.1} & 5.8 \\
\midrule
\multicolumn{10}{@{}l}{\emph{SMPL-based methods}} \\
KIMODO~\cite{kimodo}
  & 0.365 & 0.500 & 0.582 & 117.0 & 21.41 & 25.36
  & 3.54 & \textbf{2.7} & 6.3 \\
ViMoGen~\cite{vimogen}
  & 0.429 & 0.569 & 0.652 & 152.2 & 21.07 & 24.18
  & 6.95 & 23.7 & 4.4 \\
DART~\cite{dartcontrol}
  & 0.548 & 0.725 & 0.794 & 127.8 & 18.53 & 26.26
  & 3.30 & 5.9 & 4.5 \\
HY-Motion-1B~\cite{hymotion}
  & \textbf{0.785} & \textbf{0.917} & \textbf{0.951} & 13.8 & \textbf{14.82} & 27.43
  & 3.08 & 6.6 & 5.4 \\
MotionStreamer~\cite{motionstreamer}
  & 0.630 & 0.787 & 0.850 & \underline{12.2} & 16.58 & \textbf{27.46}
  & 5.00 & 17.1 & 11.1 \\
\midrule
\name (ours)
  & \underline{0.752} & \underline{0.878} & \underline{0.925} & \textbf{10.1} & \underline{15.51} & \underline{27.44}
  & 3.06 & 6.3 & 7.2 \\
\bottomrule
\end{tabular}%
    }
\end{table*}

\section{Introduction}
\label{sec:intro}
\IEEEPARstart{T}{ext}-driven 3D human motion generation must combine semantic alignment and motion fidelity with prefix control, multi-action composition, and stable long-horizon rollout. Progress has come from larger motion--text corpora~\cite{t2m,motionx,vermo,gotozero} and diffusion, flow-matching, and autoregressive generators~\cite{mdm,mld,motiondiffuse,mfm,discord,t2mgpt,momask,motiongpt}.

Most such models nevertheless generate coarse frame- or clip-level units. Discrete-token methods~\cite{t2mgpt,motiongpt,momask,mmm}, raw-feature diffusion or flow models~\cite{mdm,motiondiffuse,remodiffuse,hymotion}, and continuous latent models~\cite{mld,motionlcm,mfm,motionstreamer} generally entangle root translation, global orientation, and joint articulation before generation. The generator must then learn text semantics and recover the hidden kinematic factorization simultaneously, complicating fine-grained reconstruction and structured conditioning.

Existing structure-aware designs indicate that body granularity matters, but do not provide a continuous body-addressable state for a general generator. MCM~\cite{mcm} models inter-joint correlations through network attention while its generated state remains coarse. MoGenTS~\cite{mogents} uses a 2D VQ token map with an explicit body-joint axis but remains tied to discrete quantization and masked-token decoding. These methods expose the token-granularity problem yet leave open the state needed for flow matching, per-token conditioning, and long-horizon rollout: a continuous latent whose cells remain tied to physical body units.

Our central claim is that kinematic-unit latent factorization provides a better generation substrate than monolithic frame-level motion latents. Human motion already has such units: the root controls global trajectory, the pelvis and spine propagate body orientation, and distal joints carry fast local dynamics. Aligning latent tokens with these units changes the regression problem seen by the generator. The latent space can be regularized per unit, known and unknown regions can be assigned different noise levels token by token, and attention can operate over stable body-unit identities instead of an undifferentiated vector.

We instantiate this insight as \textbf{PRISM} (\textbf{P}e\textbf{r}-unit~Kinematic Motion Generation \textbf{i}n a \textbf{S}tructured Latent \textbf{M}anifold; \figref{fig:teaser}). PRISM first encodes motion with a continuous, causal Motion VAE whose latent grid contains one spatial token for each kinematic unit. A Kinematic-Unit Flow Transformer (KU-FlowT) then performs text-conditioned flow matching on this grid. We use \emph{structured latent manifold} to denote the resulting continuous, VAE-regularized motion space together with its explicit time and kinematic-unit coordinates: samples are generated in one connected latent space, while each coordinate remains tied to a physical body unit rather than an anonymous channel. These stable identities support per-token timestep labels, kinematic-tree rotary coordinates, and kinematic-adaptive inference schedules.

Following Diffusion Forcing~\cite{diffusionforcing}, token-wise timesteps express text-to-motion, prefix conditioning, and segment-wise rollout as one clean-context/noisy-target problem. KT-RoPE replaces arbitrary joint indices with skeleton-derived coordinates, while KAFS varies inference refinement with kinematic depth without retraining.

We evaluate one 1.4B-parameter model trained solely on public MotionHub~\cite{vermo} sources. PRISM obtains the best HumanML3D FID among evaluated methods and the strongest retrieval metrics among academic-data baselines, while remaining competitive with HY-Motion~\cite{hymotion}, which uses much more non-public data. Without task-specific retraining, it also leads the primary paired-reconstruction, prefix-conditioned, BABEL sequential, and narrative-preference comparisons; controlled ablations connect these gains to the latent layout and topology-aware operations.

Our contributions are summarized as follows:
\begin{enumerate}[leftmargin=*,itemsep=0.2ex]
\item We introduce a continuous, causal \textbf{kinematic-unit latent manifold} whose body-addressable grid improves paired reconstruction and text-conditioned generation over a matched monolithic latent.
\item We exploit this grid through \textbf{KT-RoPE} for topology-aware attention and training-free \textbf{KAFS} for depth-aware sampling, with controlled ablations validating both designs.
\item We formulate text, prefix-pose, sequential, and narrative generation as one \textbf{clean-context/noisy-target} problem. A single 1.4B-parameter SMPL-native model trained only on public academic data performs strongly across all four regimes without task-specific retraining.
\end{enumerate}

\begin{figure*}[t]
\centering
\includegraphics[width=\textwidth,keepaspectratio]{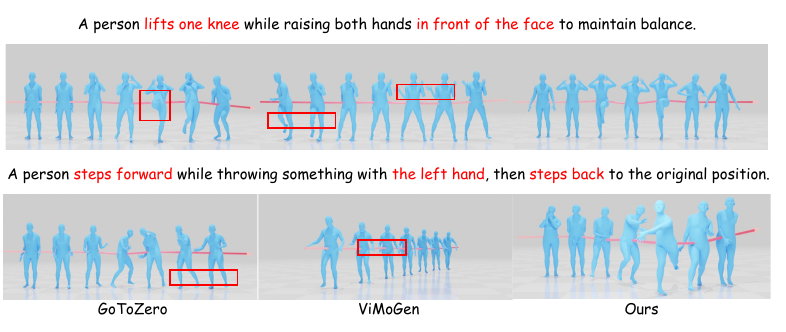}
\caption{Qualitative comparison on text-to-motion. Red prompt words mark key action constraints, and red boxes/lines highlight visible mismatches or artifacts in the compared generations.}
\label{fig:comp_t2m}
\end{figure*}

\section{Related Work}
\label{sec:related}

\subsection{Text-to-motion generation}
Text-to-motion has progressed from sequence models and VAEs~\cite{temos,cai2021unified} to raw or latent diffusion~\cite{mdm,motiondiffuse,remodiffuse,mld}, consistency models~\cite{motionlcm}, flow matching~\cite{mfm,discord}, and autoregressive or masked-token generators~\cite{t2mgpt,motiongpt,momask,mmm,motiongpt3}. Recent systems further scale data and model capacity: Go-To-Zero~\cite{gotozero} studies million-scale data, HY-Motion~\cite{hymotion} uses a billion-parameter DiT-style flow model~\cite{dit}, and ViMoGen~\cite{vimogen} studies generalizable SMPL generation. These advances improve objectives and scale but do not determine the granularity of the generated motion unit.

\subsection{Motion tokenization and joint structure}
The granularity of motion tokens or latents is a distinct design axis from the generative objective. Many discrete-token systems~\cite{t2mgpt,momask,motiongpt,gotozero} quantize each pose or motion segment into compact codes, which makes sequence modeling convenient but places root motion, global orientation, and local joint rotations inside a monolithic generated unit. Continuous latent models~\cite{mld,mfm,motionstreamer} avoid discrete codebooks, yet commonly use latent vectors whose coordinates are not tied to stable body units. As a result, the network may learn body coordination, but the generated token itself remains body-structure agnostic.

Several works explicitly study joint structure in motion generation. MCM~\cite{mcm} introduces multi-wise attention to model correlations among joints, and MoGenTS~\cite{mogents} argues that pose-level quantization loses spatial relationships among joints, replacing it with a joint-level spatial-temporal VQ token map. These works make the token-granularity issue visible and show that joint-aware modeling can improve motion synthesis. At the same time, existing joint-aware approaches are largely tied to either network-side attention design or discrete VQ tokenization. \name instead keeps a continuous SMPL latent grid whose cells remain body-addressable during flow matching, prefix conditioning, KT-RoPE, and KAFS.

\subsection{Conditional and long-horizon motion generation}
Conditional methods use diffusion inpainting~\cite{priormdm}, blended positional encodings~\cite{flowmdm}, action composition~\cite{temos}, or explicit poses~\cite{kimodo}; streaming systems such as MotionStreamer~\cite{motionstreamer}, AMD~\cite{amd}, and ScaMo~\cite{scamo} address generated history. Diffusion Forcing~\cite{diffusionforcing} provides token-specific noise conditioning more generally. Unified systems span multimodal token or language-model interfaces~\cite{ude,lmm,unimumo,motiongpt,motiongpt2,motiongpt3,motionagent,lom,motioncraft} and common generation/editing interfaces~\cite{motionlab,vermo}. PRISM changes a complementary axis: prefix frames, generated history, and targets share one clean-context/noisy-target interface whose latent coordinates remain physical body units.

\begin{table*}[t]
\caption{Prefix-pose-conditioned generation on HumanML3D.}
\label{tab:tp2m}
\centering
\begingroup
\setlength{\tabcolsep}{3.5pt}
\renewcommand{\arraystretch}{1.05}
\begin{tabular*}{\textwidth}{@{\extracolsep{\fill}} c l cccccc ccc @{}}
\toprule
\multirow{2}{*}{Prefix}
  & \multirow{2}{*}{Method}
  & \multicolumn{6}{c}{HumanML3D~\cite{t2m}}
  & \multicolumn{3}{c}{Physical Plausibility} \\
\cmidrule(lr){3-8} \cmidrule(lr){9-11}
  & & R-P~T1$\uparrow$ & R-P~T2$\uparrow$ & R-P~T3$\uparrow$ & FID$\downarrow$ & MM~D$\downarrow$ & Div.$\rightarrow$
  & Slide$\downarrow$ & Float$\downarrow$ & Jitter$\downarrow$ \\
\midrule
-- & Real motion
  & 0.770 & 0.903 & 0.944 & 0.0 & 14.88 & 27.77 & 2.00 & 2.6 & 5.5 \\
\midrule
\multirow{5}{*}{1 frame}
  & KIMODO~\cite{kimodo}        & 0.525 & 0.690 & 0.769 & 113.0 & 21.06 & 25.97 & \underline{2.20} & \textbf{3.4} & \underline{5.4} \\
  & FlowMDM~\cite{flowmdm}      & 0.449 & 0.630 & 0.706 & 89.9 & 19.87 & 25.77 & 3.72 & 8.0 & 6.2 \\
  & MotionLab~\cite{motionlab}  & 0.526 & 0.671 & 0.744 & 58.3 & 18.79 & \textbf{27.11} & \textbf{1.97} & \underline{3.6} & \textbf{5.0} \\
  & MotionStreamer~\cite{motionstreamer} & \underline{0.617} & \underline{0.780} & \underline{0.850} & \underline{14.7} & \underline{16.93} & 27.00 & 5.67 & 18.4 & 11.5 \\
  & \name (ours)                & \textbf{0.780} & \textbf{0.904} & \textbf{0.938} & \textbf{10.2} & \textbf{15.20} & \underline{27.41} & 3.48 & 9.8 & 8.6 \\
\midrule
\multirow{5}{*}{5 frames}
  & KIMODO~\cite{kimodo}        & 0.538 & 0.699 & 0.775 & 109.4 & 20.96 & 25.93 & \underline{2.05} & \underline{4.1} & \underline{5.0} \\
  & FlowMDM~\cite{flowmdm}      & 0.481 & 0.654 & 0.729 & 82.2 & 19.49 & 26.66 & 3.35 & 8.0 & 6.4 \\
  & MotionLab~\cite{motionlab}  & 0.502 & 0.639 & 0.720 & 67.6 & 19.32 & \textbf{27.04} & \textbf{2.00} & \textbf{3.8} & \textbf{4.7} \\
  & MotionStreamer~\cite{motionstreamer} & \underline{0.628} & \underline{0.786} & \underline{0.853} & \underline{12.8} & \underline{16.78} & \underline{26.98} & 5.26 & 18.4 & 12.5 \\
  & \name (ours)                & \textbf{0.778} & \textbf{0.905} & \textbf{0.943} & \textbf{9.4} & \textbf{15.11} & \textbf{27.38} & 3.20 & 8.5 & 8.4 \\
\midrule
\multirow{5}{*}{9 frames}
  & KIMODO~\cite{kimodo}        & 0.531 & 0.704 & 0.772 & 109.4 & 20.96 & 25.92 & \textbf{2.01} & \underline{4.4} & \underline{4.8} \\
  & FlowMDM~\cite{flowmdm}      & 0.490 & 0.664 & 0.742 & 77.8 & 19.31 & 26.38 & 3.44 & 8.0 & 6.3 \\
  & MotionLab~\cite{motionlab}  & 0.502 & 0.649 & 0.722 & 70.5 & 19.35 & \underline{26.92} & \underline{2.05} & \textbf{4.0} & \textbf{4.7} \\
  & MotionStreamer~\cite{motionstreamer} & \underline{0.633} & \underline{0.788} & \underline{0.856} & \underline{12.9} & \underline{16.64} & \textbf{27.36} & 5.25 & 16.9 & 12.3 \\
  & \name (ours)                & \textbf{0.793} & \textbf{0.909} & \textbf{0.940} & \textbf{9.1} & \textbf{15.05} & \underline{27.41} & 3.20 & 8.1 & 8.3 \\
\bottomrule
\end{tabular*}
\endgroup
\end{table*}

\begin{table*}[t]
\caption{Sequential action generation on official BABEL validation episodes.}
\label{tab:babel_seq}
\centering
\setlength{\tabcolsep}{3.2pt}
\scriptsize
\renewcommand{\arraystretch}{1.05}
\resizebox{\textwidth}{!}{%
\begin{tabular}{@{}l cccc cc cc c@{}}
\toprule
\multirow{2}{*}{Method}
  & \multicolumn{4}{c}{Full BABEL}
  & \multicolumn{5}{c}{Long-Horizon Stress Slices} \\
\cmidrule(lr){2-5} \cmidrule(lr){6-10}
  & R@3$\uparrow$ & FID$\downarrow$ & MM~D$\downarrow$ & T-FID$\downarrow$
  & Long R@3$\uparrow$ & Long FID$\downarrow$
  & Act. R@3$\uparrow$ & Act. FID$\downarrow$
  & Trans. FID$\downarrow$ \\
\midrule
Real
  & 0.653 & 0.0 & 19.73 & 0.0
  & 0.632 & 0.0 & 0.621 & 0.0 & 0.0 \\
\midrule
DoubleTake~\cite{priormdm}
  & 0.448 & 63.4 & 22.47 & \underline{61.4}
  & 0.368 & 60.5 & 0.343 & 62.7 & \underline{55.8} \\
FlowMDM~\cite{flowmdm}
  & \underline{0.522} & \underline{29.3} & \underline{21.04} & \textbf{33.5}
  & \underline{0.485} & \underline{39.4} & \underline{0.473} & \underline{36.4} & \underline{40.1} \\
MotionStreamer~\cite{motionstreamer}
  & 0.397 & 51.3 & 22.21 & 68.0
  & 0.321 & 64.1 & 0.308 & 59.2 & 75.1 \\
\name (ours)
  & \textbf{0.625} & \textbf{28.6} & \textbf{20.09} & \underline{38.9}
  & \textbf{0.614} & \textbf{26.6} & \textbf{0.591} & \textbf{29.0} & \textbf{36.8} \\
\bottomrule
\end{tabular}%
}
\end{table*}

\section{Method}
\label{sec:method}

\subsection{Overview}
\label{sec:method_overview}

\name addresses text-driven SMPL motion generation by changing the unit of generation. Instead of denoising a holistic per-frame vector or a sequence-level latent, it learns a continuous latent grid whose coordinates follow time and kinematic-unit axes. Each latent token therefore keeps a stable physical meaning: the generator can mark observed tokens as clean context and denoise target tokens, inject kinematic-tree structure along the joint axis, and apply joint-adaptive sampling within one architecture.

As illustrated in \figref{fig:pipeline}, the framework first represents each SMPL frame as physically meaningful tokens: root translation, global orientation, and local joint rotations (\secref{sec:motion-representation}).  A causal Motion VAE maps this motion grid to a continuous latent grid that preserves the time--kinematic-unit layout and decodes back to SMPL parameters (\secref{sec:motion-vae}).

Generation is performed by a \emph{Kinematic-Unit Flow Transformer} (KU-FlowT), a latent flow-matching Transformer that predicts missing target tokens under text conditioning. Because the spatial axis of the latent grid has stable body meanings, KU-FlowT can use per-token timesteps, topology-aware RoPE coordinates, and depth-aware inference schedules on the same representation (\secref{sec:flow-transformer}). The clean-context/noisy-target interface is then reused for autoregressive segment chaining and narrative motion composition (\secref{sec:autoregressive_composition}).

\subsection{Motion Representation}
\label{sec:motion-representation}

PRISM uses a kinematic-unit motion representation rather than a holistic per-frame vector.  For a sequence of $T$ frames, SMPL~\cite{smpl} provides a root translation $\mathbf{p}_i\in\mathbb{R}^{3}$, a global orientation $\boldsymbol{\theta}_i^0\in\mathbb{R}^{6}$, and local joint rotations $\{\boldsymbol{\theta}_i^j\in\mathbb{R}^{6}\}_{j=1}^{J}$ in the 6D rotation representation~\cite{rot6d}.  We arrange these quantities as $K=J+2$ spatial tokens per frame:
\begin{equation}
\label{eq:joint_grid}
X = \bigl[\,\underbrace{[\mathbf{p}_i;\,\Delta\mathbf{p}_i]}_{\text{root}},\;
\underbrace{\boldsymbol{\theta}_i^0}_{\text{orientation}},\;
\underbrace{\boldsymbol{\theta}_i^1,\ldots,\boldsymbol{\theta}_i^J}_{\text{joint rotations}}\,\bigr]_{i=1}^{T}
\in\mathbb{R}^{T\times K\times 6},
\end{equation}
where $\Delta\mathbf{p}_i=\mathbf{p}_i-\mathbf{p}_{i-1}$.  This representation is the interface used by the entire method: it gives the VAE separately regularized physical units, gives the Transformer a meaningful joint axis, and makes clean/noisy token assignment well-defined during conditional generation.

\subsection{Causal Kinematic-Unit Motion VAE}
\label{sec:motion-vae}

The Motion VAE maps the kinematic-unit motion grid $X$ to a latent grid $Z$ while preserving the time--kinematic-unit structure needed by the generator.  We refer to this representation as a \emph{structured latent manifold}: a continuous, VAE-regularized motion space whose coordinates form an explicit time$\times$kinematic-unit grid.  The term highlights two properties used by the generator.  First, flow matching transports Gaussian samples within one continuous latent space rather than selecting from a discrete VQ codebook.  Second, each coordinate remains bound to one physical body unit (root translation, global orientation, or a specific joint) rather than an anonymous channel.  The encoder applies causal temporal convolutions across multiple resolutions and performs spatial joint interaction at the bottleneck by attending across the $K$ kinematic units at each latent time step; the decoder mirrors this hierarchy and outputs SMPL-compatible rotations and root motion.  The shared legend in \figref{fig:pipeline} identifies these standard operators, while block widths indicate relative temporal scale rather than literal token counts.  Causal layers reuse cached feature tails across chunks without future leakage; this causal design also allows the final frames of one generated segment to be encoded and reused as the clean prefix of the next.  Body-addressability is equally important: each latent spatial position has a stable body meaning, so downstream attention, timestep conditioning, and kinematic schedules can operate on physical units rather than anonymous dimensions.

The VAE is trained to reconstruct both the SMPL parameters and their kinematic consequences:
\begin{equation}
\label{eq:vae_loss}
\mathcal{L}_{\text{VAE}} =
\lambda_{\text{p}}\mathcal{L}_{\text{p}}
+\lambda_{\text{j}}\mathcal{L}_{\text{j}}
+\lambda_{\text{tr}}\mathcal{L}_{\text{tr}}
+\lambda_{\text{KL}}\mathcal{L}_{\text{KL}}.
\end{equation}
Here $\mathcal{L}_{\text{p}}$ reconstructs root motion and rotations, and $\mathcal{L}_{\text{KL}}$ regularizes the posterior.  Because local rotation errors can accumulate along the body chain, $\mathcal{L}_{\text{j}}$ applies forward kinematics to the reconstructed rotations and penalizes root-relative joint positions.  The trajectory term supervises the rollout induced by predicted displacements,
\begin{equation}
\label{eq:traj_loss}
\mathcal{L}_{\text{tr}} =
\frac{1}{T}\sum_{i=1}^{T}
\Bigl\|\sum_{k=1}^{i}\Delta\hat{\mathbf{p}}_k
- \sum_{k=1}^{i}\Delta\mathbf{p}_k\Bigr\|_1,
\end{equation}
so early displacement errors are penalized through their effect on all later root positions.  This objective keeps the latent grid close to deployment-ready SMPL motion rather than merely reconstructing an evaluation skeleton.

\begin{table*}[t]
\caption{Paired motion reconstruction on HumanML3D.}
\label{tab:paired_recon_cmp}
\centering
\setlength{\tabcolsep}{4.0pt}
\tiny
\renewcommand{\arraystretch}{1.08}
\resizebox{\textwidth}{!}{%
\begin{tabular}{@{}l cccccccc c@{}}
\toprule
\multirow{2}{*}{Method}
  & \multicolumn{8}{c}{Reconstruction (HumanML3D)}
  & \multirow{2}{*}{ICR} \\
\cmidrule(lr){2-9}
  & MPJPE$\downarrow$ & RA-MPJPE$\downarrow$ & dRoot-XZ$\downarrow$ & dRoot-Y$\downarrow$ & PA-MPJPE$\downarrow$ & MPJRE$\downarrow$ & rFID$\downarrow$ & Emb-L2$\downarrow$
  & \\
\midrule
MLD~\cite{mld,motionlcm}
  & 118.54 & 44.29 & 3.61 & 2.41 & 33.62 & 8.43 & 53.19 & 13.32 & -- \\
MoGenTS~\cite{mogents}
  & 104.42 & 32.17 & 1.81 & \underline{0.90} & 20.99 & 7.19 & 46.07 & 11.52 & $0.16{\times}$ \\
MotionStreamer~\cite{motionstreamer}
  & \underline{40.02} & \underline{25.88} & \underline{0.59} & 1.73 & \underline{19.18} & \underline{5.21} & \underline{2.08} & \underline{4.19} & $1.06{\times}$ \\
\midrule
\name (ours)
  & \textbf{9.72} & \textbf{1.74} & \textbf{0.50} & \textbf{0.24} & \textbf{1.09} & \textbf{0.30} & \textbf{1.04} & \textbf{1.51} & $1.5{\times}$ \\
\bottomrule
\end{tabular}%
}
\end{table*}

\subsection{Kinematic-Unit Flow Transformer}
\label{sec:flow-transformer}

KU-FlowT is the PRISM generator over the VAE latent grid.  Given the encoded text condition $c$, clean context tokens, and noisy target tokens, it predicts the flow-matching velocity field that transports noise toward the clean latent motion.  The generator adapts a DiT-style Transformer to flattened time--kinematic-unit latent tokens with text cross-attention.

The kinematic-unit grid gives KU-FlowT three controls that are unavailable in a monolithic latent. Per-token timestep conditioning defines which tokens are observed or generated, KT-RoPE supplies skeleton-derived joint-axis positions to the standard RoPE map, and KAFS reparameterizes the inference schedule per joint. The key difference from monolithic latent generation is that the model denoises a grid whose spatial positions already correspond to body units.

\subsubsection{Per-token timestep conditioning}
\label{para:nfci}

Per-token timestep conditioning implements the clean-context/noisy-target interface introduced above.  Let $\Omega_{\mathrm{ctx}}$ be the tokens belonging to the prefix and $\Omega_{\mathrm{tar}}$ the tokens to generate.  For each token $r$, we assign a timestep $\tau_r$:
\begin{equation}
\label{eq:problem_mixed_state}
Z_{\tau,r}=(1-\tau_r)Z_{0,r}+\tau_r\epsilon_r,\qquad
\epsilon_r\sim\mathcal{N}(0,I),
\end{equation}
where $\tau_r=0$ for clean context tokens and $\tau_r>0$ for target tokens.  The Transformer therefore learns
\begin{equation}
\label{eq:problem_latent}
p_{\theta}\!\left(Z_{\Omega_{\mathrm{tar}}}\mid c, Z_{\Omega_{\mathrm{ctx}}}\right)
\end{equation}
without changing architecture across modes.  Text-to-motion is the all-target case, prefix-pose generation keeps the prefix clean, and autoregressive generation reuses the previous segment tail as clean context.

\subsubsection{Kinematic-Tree Rotary Position Embeddings}
\label{para:kt_rope}

Standard joint-axis RoPE~\cite{rope} indexes tokens by the arbitrary body-joint storage order, so kinematically adjacent joints can receive distant positions while unrelated joints can appear adjacent.  KT-RoPE replaces this storage coordinate with a scalar coordinate derived from the kinematic tree while preserving the standard scalar RoPE interface.  Let $G=(\mathcal{V},\mathcal{E})$ be the kinematic tree with adjacency $A$, degree matrix $D$, and normalized graph Laplacian
\begin{equation}
\label{eq:kt_laplacian}
\mathcal{L}=I-D^{-1/2}AD^{-1/2}.
\end{equation}
Using the first non-trivial eigenvectors of $\mathcal{L}$, each joint obtains a spectral coordinate
\begin{equation}
\label{eq:spectral_coord}
\mathbf{c}_j=[u_1(j),\ldots,u_r(j)].
\end{equation}
We project it to one signed residual and add it to the pretrained sequential body-token coordinate,
\begin{equation}
\label{eq:kt_scalar_projection}
\begin{aligned}
q_j&=\mathbf{w}^{\top}\mathbf{c}_j+\epsilon\pi_{\mathrm{DFS}}(j),\\
r_j&=\frac{q_j-\bar{q}}{\max_{\ell}|q_{\ell}-\bar{q}|},\qquad
p_j=b_j+\lambda r_j,
\end{aligned}
\end{equation}
where $b_j$ is the original body-token coordinate, $\pi_{\mathrm{DFS}}$ is a deterministic depth-first tie-breaker, and $\lambda$ controls the topology residual.  The translation token uses the reserved position 0 with identity RoPE.  The joint-axis rotary map then uses the usual frequency ladder,
\begin{equation}
\label{eq:kt_rope_freq}
R_j^{(i)} = R(p_j\theta_i).
\end{equation}
Thus topology changes only the joint coordinate $p_j$; attention dimensions, channel pairing, and the RoPE frequency basis itself remain unchanged.

\subsubsection{Kinematic-Adaptive Flow Scheduling}
\label{para:kafs}

KAFS is the sampler-side counterpart of KT-RoPE: both use the physical address of each latent token.  Per-token noise levels are valid conditioning variables in Diffusion Forcing~\cite{diffusionforcing}, diffusion samplers are sensitive to explicit time/noise schedule design~\cite{karras2022edm}, and human motion exhibits proximal-to-distal coordination along the kinematic chain~\cite{serrien2017proximal}.  KAFS therefore maps kinematic depth to flow time: root and spine tokens retain the base schedule that controls global trajectory and orientation, while deeper limb tokens receive progressively shifted schedules for finer late-stage articulation.  It keeps the same Euler step index $k$ as the baseline sampler and only warps the noise level for each body token:
\begin{equation}
\label{eq:kafs}
\begin{aligned}
\sigma_j^{(k)} &= \bigl(\sigma^{(k)}\bigr)^{\gamma_j},
\qquad
\gamma_j=\frac{1}{\alpha_j},\\
\alpha_j&=1-\frac{d_j}{d_{\max}}(1-\alpha_{\min}),
\end{aligned}
\end{equation}
where $d_j$ is kinematic depth.  We instantiate the schedule with $\alpha_{\min}{=}0.85$, i.e., a $0.025$ decrement per depth band for the SMPL kinematic tree used in our latent grid.  This gives the deepest end-effectors a maximum exponent of $1/0.85{\approx}1.18$ while leaving root-level tokens on the original flow schedule.  At step $k$, the token-specific noise level $\sigma_j^{(k)}$ is used both as the timestep label passed to the network and as the noise level that determines the Euler step size:
\begin{equation}
\label{eq:kafs_step}
x_j^{(k+1)}\leftarrow x_j^{(k)}+
\bigl(\sigma_j^{(k+1)}-\sigma_j^{(k)}\bigr)
v_{\theta,j}\!\bigl(x^{(k)},\{\sigma_{j'}^{(k)}\}_{j'},c\bigr).
\end{equation}
Because the timestep label and Euler increment are derived from the same monotone schedule, KAFS changes only the sampling trajectory; it adds no parameters and does not change the trained model.

\subsection{Autoregressive and Narrative Motion Composition}
\label{sec:autoregressive_composition}

Segment-wise autoregressive generation repeatedly applies the clean-context/noisy-target latent generator in Eq.~\eqref{eq:problem_latent}.  The first segment is generated from text or an optional user prefix; the last $F$ frames are then VAE-encoded and injected as clean tokens for the next segment.  This produces long motions without post-hoc blending because the boundary condition is part of the denoising process itself.

For free-form narrative prompts, we follow HY-Motion-style text normalization~\cite{hymotion} and prior segment-based long-horizon systems~\cite{motionstreamer,flowmdm,priormdm} to decompose the narrative into atomic action prompts with predicted durations.  Each atomic prompt is encoded as one segment condition $c_n$, and PRISM composes the full motion by the same autoregressive chaining mechanism.  Narrative composition is therefore not a separate model; it is an application of the shared clean-context/noisy-target formulation to a prompt sequence.

\begin{figure*}[t]
\centering
\includegraphics[width=\textwidth,keepaspectratio]{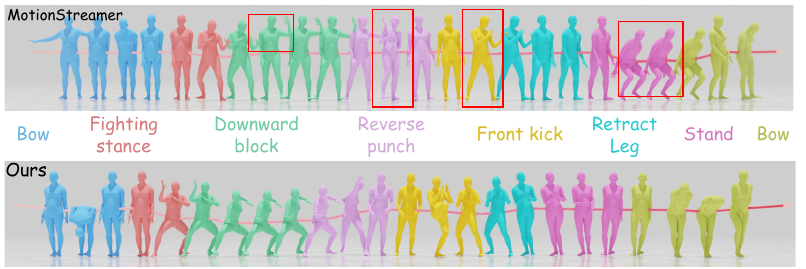}
\caption{Qualitative comparison on long-horizon narrative composition. Colors denote consecutive atomic action segments from the shared segment plan, and red boxes highlight missed or unnatural actions in the baseline sequence.}
\label{fig:comp_long}
\end{figure*}

\section{Experiments}
\label{sec:experiments}

We evaluate \name through four claims: public-data text-to-motion with animatable SMPL output, clean-prefix conditioning without task-specific retraining, long-horizon segment reuse, and the contribution of the proposed latent and topology design. The experiments proceed from generation benchmarks to reconstruction and ablations, separating representation quality from generator scale.

\subsection{Experimental Setup}
\label{sec:setup}

\subsubsection{Datasets and benchmarks}
\name is trained on ${\sim}200$K curated SMPL(-X) motion--text pairs drawn entirely from publicly available MotionHub~\cite{vermo} sources. A single model is evaluated on four settings: \emph{text-to-motion} (T2M) and \emph{prefix-pose-conditioned generation} (TP2M) on HumanML3D~\cite{t2m}, where TP2M conditions on $1$, $5$, or $9$ clean initial frames; \emph{long-horizon sequential generation} under the BABEL~\cite{babel} protocol of FlowMDM~\cite{flowmdm}; and \emph{narrative motion composition} on $50$ free-form multi-action prompts spanning $10$ categories ($4$--$10$ sub-actions each). For the unified-generation experiments, the trained generator, VAE, text encoder, and sampler are fixed; T2M uses an all-target mask, TP2M marks the prefix frames as clean context, and sequential/narrative generation reuses the previous segment tail as clean context for the next segment.

\subsubsection{Model and inference configuration}
All experiments use the same Motion VAE and KU-FlowT unless an ablation explicitly states otherwise. The Motion VAE uses causal temporal convolutions with $4{\times}$ temporal downsampling and represents each frame as $K{=}23$ kinematic-unit tokens: root motion, global orientation, and $21$ local joint rotations in the 6D rotation representation~\cite{rot6d}. Each token has latent dimension $D{=}16$, and the VAE is trained for $400$K steps on $16$ V100 GPUs with batch size $64$ per GPU.

KU-FlowT follows a DiT-style backbone~\cite{dit,wan} with $30$ blocks, T5-XXL text conditioning~\cite{t5}, KT-RoPE joint-axis coordinates, and ${\sim}1.4$B parameters. It is trained for $300$K steps on $80$ V100 GPUs with batch size $6$ per GPU. At inference, we use $50$ Euler steps with classifier-free guidance scale $5.0$. The default model uses projected-spectral KT-RoPE and depth-driven KAFS.

\subsubsection{Baselines and shared SMPL evaluation}
\name generates SMPL parameters directly, whereas many HumanML3D baselines output joint-feature motion. We use ``SMPL-based'' to denote methods that directly produce SMPL-family coefficients. For KIMODO~\cite{kimodo}, we use its released SMPL-X model so that the comparison does not introduce additional inverse-kinematics degradation. For all methods, generated motions are first converted to SMPL-family coefficients when needed, and are then processed into the shared representation used by the TMR evaluator.

\subsubsection{Metrics}
T2M and TP2M report R-Precision, FID, MM-Dist, and Diversity with a TMR evaluator, while BABEL reports R@3, FID, MM-Dist, and transition FID over stitched action segments. Narrative composition uses a tie-aware ranking study: for each text input, evaluators rank anonymous videos from the compared methods, with ties allowed when outputs are not reliably separable. Where applicable in generation tables, we further report MBench-style physical-plausibility errors~\cite{vimogen} recomputed on a shared SMPL-22 FK skeleton after aligning each clip's lowest foot to the ground. In all tables, \textbf{bold} marks the best and \underline{underline} the second best generated result for that metric, ``--'' denotes that a method does not provide a same-protocol measurement, and arrows give metric direction ($\uparrow$ higher, $\downarrow$ lower, $\rightarrow$ closer to Real is better).

\subsection{Text-to-Motion}
\label{sec:t2m}

\tabref{tab:eval_t2m} evaluates whether animatable SMPL output can retain feature-space text-to-motion quality, and \figref{fig:comp_t2m} visualizes representative failures. The most informative comparisons are structurally close methods: ViMoGen also uses a Transformer-style SMPL generator but has lower R-Precision and higher FID than \name; DART and KIMODO output SMPL-family motion but trail \name on retrieval; and MLD/MotionLCM have higher FID and MM-Dist, suggesting that generic latent compression is not equivalent to body-addressable latent modeling. HY-Motion remains stronger on R-Precision and MM-Dist with large-scale non-public motion data, whereas \name obtains lower FID using only public academic data.

\begin{table}[!t]
\caption{Kinematic-unit-factorized versus monolithic latent spaces under controlled reconstruction and generation settings.}
\label{tab:abl_2d1d}
\centering
\scriptsize
\renewcommand{\arraystretch}{1.05}
\setlength{\tabcolsep}{3pt}
\resizebox{\columnwidth}{!}{%
\begin{tabular}{@{}l cc ccc@{}}
\toprule
\multirow{2}{*}{Latent}
  & \multicolumn{2}{c}{Recon.}
  & \multicolumn{3}{c}{T2M Gen.} \\
\cmidrule(lr){2-3}\cmidrule(lr){4-6}
  & rFID$\downarrow$ & MPJPE$\downarrow$
  & FID$\downarrow$ & R-P~T3$\uparrow$ & MM-D$\downarrow$ \\
\midrule
1D (monolithic)   & 8.90 & 45.13 & 36.7 & 0.822 & 17.08 \\
2D (joint, ours)  & \textbf{1.04} & \textbf{9.72} & \textbf{10.1} & \textbf{0.925} & \textbf{15.51} \\
\bottomrule
\end{tabular}%
}
\end{table}

\subsection{Pose-Conditioned Generation}
\label{sec:tp2m}

\tabref{tab:tp2m} tests the clean-context/noisy-target interface rather than a separately trained pose-conditioned model. Using the T2M-trained generator without task-specific retraining, \name gives the best R-Precision, FID, and MM-Dist for all prefix lengths; FID decreases from 10.2 with a one-frame prefix to 9.1 with a nine-frame prefix, and MM-Dist from 15.20 to 15.05. Thus prefix frames can be kept clean while the target region is generated under text conditioning. MotionLab and KIMODO are smoother on selected contact metrics, but they do not match PRISM's prompt-prefix semantic scores.

\subsection{Long-Horizon Sequential Generation}
\label{sec:sequential}

\tabref{tab:babel_seq} asks whether the prefix interface remains useful when generation is repeated over action segments. On full BABEL episodes, \name improves R@3 over FlowMDM from 0.522 to 0.625 and slightly improves FID/MM-Dist. The stress slices are more diagnostic: \name obtains the best long-clip and many-action R@3/FID, showing that semantic recoverability does not collapse as segment count increases. The main trade-off is transition distribution, where FlowMDM's BABEL-specific design keeps the best full-set T-FID; PRISM instead provides one interface that transfers from T2M and TP2M to sequential rollout.

\subsection{Paired Motion Reconstruction}
\label{sec:paired_recon}

We next isolate the motion representation because all downstream generation is performed in the VAE latent space. \tabref{tab:paired_recon_cmp} evaluates all methods after conversion to the same motion135/SMPL-H forward-kinematics skeleton, using paired joint, root-displacement, rotation-direction, rFID, and embedding-distance errors. The table also reports the information compression rate (ICR), computed as the number of input feature scalars divided by the decoder-side latent/feature tensor size after temporal downsampling, so reconstruction quality can be interpreted together with how much motion information each tokenizer retains. Values below $1{\times}$ indicate an expanded latent tensor, and ``--'' denotes a clip-level latent whose ICR varies with sequence length.

\name records the lowest errors across geometry and TMR feature metrics, which is the direct evidence that the latent grid preserves articulated structure before text generation is considered. Compared with MotionStreamer, MPJPE drops from 40.02 mm to 9.72 mm, RA-MPJPE from 25.88 mm to 1.74 mm, MPJRE from 5.21$^\circ$ to 0.30$^\circ$, and rFID from 2.08 to 1.04. The ICR column also shows that reconstruction quality is not determined by latent tensor size alone: MoGenTS uses an expanded dual-stream latent tensor, and MotionStreamer uses a high-channel causal feature map after temporal downsampling, yet both have higher reconstruction errors. This establishes \name as a stronger motion tokenizer/reconstructor than the compared baselines, while the matched 1D/2D ablation below tests whether the gain comes from the kinematic-unit-factorized layout and transfers to text-conditioned generation.

\paragraph*{Reconstruction vs.\ generation headroom}
Recent latent-generation work shows that stronger reconstruction does not automatically improve generation, because a more detailed latent space can be harder to model~\cite{vavae}. We therefore use \tabref{tab:paired_recon_cmp} as representation evidence and rely on the matched latent-layout ablation below to test whether the kinematic-unit-factorized layout is also generation-friendly.

\subsection{Ablation Studies}
\label{sec:ablations}

We ablate the latent layout, topology-aware positional encoding, and inference schedule.

\begin{table}[!t]
\caption{Kinematic-adaptive flow scheduling ablation on HumanML3D with the TMR evaluator.}
\label{tab:abl_kafs}
\centering
\scriptsize
\renewcommand{\arraystretch}{1.05}
\setlength{\tabcolsep}{3pt}
\resizebox{\columnwidth}{!}{%
\begin{tabular}{@{}l cccc@{}}
\toprule
KAFS Mode
  & FID$\downarrow$ & R-P~T3$\uparrow$ & MM-D$\downarrow$ & Div.$\rightarrow$ \\
\midrule
None (baseline)          & 12.1 & \underline{0.925} & \underline{15.47} & 27.35 \\
Uniform ($\alpha{=}1.0$) & \underline{11.0} & \textbf{0.926} & \textbf{15.46} & \underline{27.36} \\
Random ($\alpha{\sim}\mathcal{U}$) & 11.4 & 0.914 & 15.54 & 27.22 \\
Depth-driven (ours)      & \textbf{10.1} & \underline{0.925} & 15.51 & \textbf{27.44} \\
\bottomrule
\end{tabular}%
}
\end{table}

\begin{table*}[t]
\caption{RoPE design comparison.}
\label{tab:abl_rope_kt}
\centering
\scriptsize
\renewcommand{\arraystretch}{1.0}
\setlength{\tabcolsep}{3pt}
\resizebox{\textwidth}{!}{%
\begin{tabular}{@{}l cccccc cccc@{}}
\toprule
\multirow{2}{*}{Setting}
  & \multicolumn{6}{c}{HumanML3D~\cite{t2m}}
  & \multicolumn{4}{c}{BABEL Seq} \\
\cmidrule(lr){2-7}\cmidrule(lr){8-11}
  & R-P~T1$\uparrow$ & R-P~T2$\uparrow$ & R-P~T3$\uparrow$
  & FID$\downarrow$ & MM-D$\downarrow$ & Div.$\rightarrow$
  & R@3$\uparrow$ & FID$\downarrow$ & MM-D$\downarrow$ & T-FID$\downarrow$ \\
\midrule
GT         & 0.770 & 0.903 & 0.944 & 0.0 & 14.88 & 27.77 & 0.653 & 0.0 & 19.73 & 0.0 \\
\midrule
2D RoPE    & 0.730 & 0.855 & 0.908 & 12.9 & 16.70 & \underline{27.33} & 0.598 & 45.8 & 21.27 & 43.6 \\
SMPL Index & \underline{0.739} & \underline{0.861} & \underline{0.914} & \underline{12.4} & \underline{16.08} & 26.94 & \underline{0.605} & \underline{37.3} & \underline{20.64} & \underline{40.9} \\
KT-RoPE    & \textbf{0.752} & \textbf{0.878} & \textbf{0.925} & \textbf{10.1} & \textbf{15.51} & \textbf{27.44} & \textbf{0.625} & \textbf{28.6} & \textbf{20.09} & \textbf{38.9} \\
\bottomrule
\end{tabular}%
}
\end{table*}

\subsubsection{Kinematic-unit-factorized (2D) vs.\ monolithic (1D) latent space}
The controlled 1D/2D comparison in \tabref{tab:abl_2d1d} addresses the remaining question left open by the reconstruction study: whether the gains come from the kinematic-unit-tokenized latent architecture rather than from VAE strength alone, and whether those gains survive downstream generation. We therefore compare a monolithic 1D latent with the proposed kinematic-unit-factorized 2D latent while keeping all other settings identical, including data, model capacity, training schedule, text encoder, sampler, and evaluator. The kinematic-unit-factorized latent reduces reconstruction rFID from 8.90 to 1.04 and MPJPE from 45.13 mm to 9.72 mm, and the same change lowers T2M FID from 36.7 to 10.1 while improving R-P~T3 and MM-Dist. Thus, the ablation supports the intended mechanism: the time--kinematic-unit layout is not only a better autoencoding format, but also creates a latent manifold that is easier for the text-conditioned generator to model.

\subsubsection{Kinematic-Adaptive Flow Scheduling (KAFS)}
\label{sec:kafs_ablation}
\tabref{tab:abl_kafs} evaluates whether body addresses are useful at sampling time. Depth-driven KAFS gives the lowest FID and closest diversity, while uniform scaling slightly favors R-P~T3 and MM-Dist. This mixed pattern indicates that KAFS is not a universal metric booster; it is a training-free prior that shifts refinement effort toward distributional quality while leaving semantic retrieval largely controlled by the generator and latent space.

\subsubsection{Kinematic-Tree RoPE (KT-RoPE) ablation}
\label{sec:kt_rope_ablation}
\tabref{tab:abl_rope_kt} evaluates whether the joint axis should carry body topology rather than an arbitrary storage order. Moving from generic 2D RoPE to the SMPL index improves both HumanML3D and BABEL, and KT-RoPE improves every reported retrieval, FID, MM-Dist, and transition-FID column further. Thus the gain is not merely from adding a second positional axis; the coordinate assigned to that axis should reflect kinematic structure.

The original joint index has only $\rho{=}0.397$ Spearman correlation with kinematic-tree distance, whereas projected spectral KT-RoPE injects topology without changing the network architecture.

\subsection{Narrative Motion Composition}
\label{sec:narrative}

Narrative motion composition evaluates generation under a shared paragraph-derived segment plan: a frozen planner converts each prompt into ordered captions with durations, and every method generates from the same plan. The task therefore stresses open-ended segment realization and history-aware continuity rather than method-specific parsing.

The study uses $50$ prompts spanning $10$ categories and $20$ evaluators with computer-vision or graphics experience. Each evaluator ranks anonymous, randomized \name, MotionStreamer, and KIMODO triplets for motion quality, text fidelity, transition smoothness, and overall preference, producing $1{,}000$ annotations per dimension. Ties are allowed when outputs are not reliably separable. We convert each three-way rank into \name-centered pairwise outcomes and report normalized Win/Tie/Loss percentages; significance is assessed with a two-sided sign test over non-tie outcomes at $p<0.01$. Identical viewpoints, meshes, captions, durations, and realized frame counts are used across methods.

\begin{table}[!t]
\caption{Narrative motion composition user study. W/T/L reports the percentage of pairwise win, tie, and loss votes for \name.}
\label{tab:user_study}
\centering
\scriptsize
\setlength{\tabcolsep}{2.4pt}
\renewcommand{\arraystretch}{1.08}
\begin{tabular*}{\columnwidth}{@{\extracolsep{\fill}} l ccc ccc @{}}
\toprule
 \multirow{2}{*}{Criterion}
 & \multicolumn{3}{c}{vs. KIMODO}
 & \multicolumn{3}{c}{vs. MotionStreamer} \\
\cmidrule(lr){2-4}\cmidrule(l){5-7}
 & W & T & L & W & T & L \\
\midrule
Motion quality
  & 43.7 & 36.1 & 20.2
  & 78.4 & 11.7 & 9.9 \\
Text fidelity
  & 76.8 & 9.7 & 13.5
  & 69.2 & 17.1 & 13.7 \\
Transition
  & 69.4 & 17.3 & 13.3
  & 72.4 & 15.7 & 11.9 \\
Overall
  & 73.2 & 17.9 & 8.9
  & 77.1 & 14.8 & 8.1 \\
\bottomrule
\end{tabular*}
\end{table}

\tabref{tab:user_study} evaluates prompts where the output must cover multiple ordered actions, and \figref{fig:comp_long} shows how segment-level failures accumulate over a long prompt. Since all methods share the same plan, the ranking mainly reflects generation rather than parsing. KIMODO is closest on motion quality, consistent with its 700-hour optical-mocap corpus~\cite{kimodo}. In our side-by-side inspection, however, it more often misses, weakens, or reorders instructed actions under the shared segment captions, which matches its larger losses on text fidelity, transition smoothness, and overall preference. Against MotionStreamer, \name wins all dimensions by large margins, showing that the sequential interface preserves ordered narrative content and history rather than only per-segment quality.

\section{Conclusion}
\label{sec:conclusion}
This paper shows that the unit of motion generation is a central design choice for text-driven SMPL motion synthesis. \name replaces monolithic frame- or clip-level latents with a continuous kinematic-unit latent manifold, where root motion, global orientation, and per-joint rotations remain addressable across time. Built on this representation, PRISM trains KU-FlowT with per-token timestep conditioning over clean-context/noisy-target masks, giving the base generator frame-conditioned continuation and autoregressive segment chaining abilities. The same physical addresses also support topology-aware attention through KT-RoPE and inference scheduling through KAFS.

The results indicate that kinematic-unit factorization improves more than reconstruction fidelity. The tokenizer achieves lower geometry, rotation, and feature errors than prior motion tokenizers, while the matched 1D/2D ablation shows that the same layout also improves text-conditioned generation. Trained only on public academic motion--text data, \name outperforms evaluated academic-data baselines on HumanML3D and remains competitive with systems trained on large-scale non-public motion data. Its trained generator further supports prefix-conditioned generation, BABEL sequential rollout, and narrative motion composition without task-specific retraining. These findings support kinematic-unit latent factorization as a practical substrate for general SMPL motion generation: it preserves articulated structure, exposes controllable body-time coordinates, and lets multiple generation regimes share one model interface.

\section{Limitations and Future Work}
\label{sec:limitations}
The current system is intentionally scoped to the representation and conditioning interface studied in this paper, leaving three limitations for future work.
\begin{enumerate}[leftmargin=*,label=(\roman*),itemsep=0.2ex,topsep=0.3ex]
\item \emph{Body coverage.} \name currently focuses on SMPL body and torso motion. Extending the same kinematic-unit formulation to expressive hands, face, and full SMPL-X control remains future work.
\item \emph{Token granularity.} The current body tokenizer assigns one token to each kinematic unit, but this is a simple design choice rather than an optimal decomposition. Some degrees of freedom are redundant or strongly coupled, such as the SMPL collar and shoulder joints, so future tokenizers should learn or search for more compact body units.
\item \emph{Conditioning.} We currently condition only on initial prefix frames; arbitrary frames and body parts remain future work.
\end{enumerate}


\bibliographystyle{IEEEtran}
\bibliography{example_paper}

\appendices
\section{Supplementary Implementation Details}
\label{sec:impl_appendix}
This appendix gives the exact KT-RoPE and KAFS settings summarized in the implementation details of the main paper.

\subsection{Root Translation Decoding}
As defined by the motion representation in the main paper, the root token contains both the absolute root translation $\mathbf{p}_i$ and the frame displacement $\Delta\mathbf{p}_i=\mathbf{p}_i-\mathbf{p}_{i-1}$. The VAE decoder is trained to reconstruct both channels, and generated latents are decoded into the same two translation outputs. At inference time, we assemble the final root trajectory by combining them: the vertical coordinate is taken from the decoded absolute translation, $\hat{p}_i^y$, while the horizontal trajectory is rolled out from predicted displacements,
\begin{equation}
\label{eq:root_xz_rollout}
\hat{\mathbf{p}}_i^{xz}
= \mathbf{a}^{xz} + \sum_{k=s}^{i}\Delta\hat{\mathbf{p}}_k^{xz},
\end{equation}
where $\mathbf{a}^{xz}$ is the available root anchor and $s$ is the first generated frame after that anchor. For prefix-conditioned and segment-wise generation, the anchor is the last clean prefix root; for text-only generation, it is initialized from the first decoded absolute root position and rollout starts from the second frame. This hybrid decoding uses the absolute channel for body height and ground placement, while using displacement rollout for temporally coherent horizontal motion.

\subsection{KT-RoPE coordinate}
The default KT-RoPE setting uses a projected-spectral joint coordinate. The translation token keeps identity RoPE, while the body-joint tokens use the first four non-trivial eigenvectors of the normalized kinematic-tree Laplacian~\cite{chung1997spectral}, with eigenvector signs fixed by requiring the pelvis coordinate to be positive. To preserve compatibility with the sequential-RoPE baseline, KT-RoPE keeps the full joint-axis RoPE frequency basis and changes only the scalar body-token coordinate:
\begingroup\small
\begin{equation}
\label{eq:kt_rope_impl}
\begin{split}
q_j &= [u_1(j),\ldots,u_4(j)](1.0,1.7,2.9,4.3)^{\top}
      +10^{-3}\pi_{\mathrm{DFS}}(j),\\
p_j &= \mathrm{linspace}(1,22)_j
      +0.25\frac{q_j-\bar{q}}{\max_{\ell}|q_{\ell}-\bar{q}|},
\end{split}
\end{equation}
\endgroup
with RoPE base $\theta{=}10000$ and temporal RoPE unchanged.

\subsection{KAFS schedule}
Unless otherwise stated, \name inference uses depth-driven KAFS with the same $50$ Euler steps and guidance scale $5.0$ as the baseline sampler.  The schedule follows the same kinematic-depth bands as the SMPL latent grid: translation and pelvis use $\alpha{=}1.0$, and each deeper band decreases $\alpha$ by $0.025$ until distal end-effectors reach $\alpha_{\min}{=}0.85$.  This proximal-to-distal discretization keeps global root motion on the base flow schedule while assigning progressively shifted schedules to articulated limb tokens. The $23$ KAFS coefficients follow the token order [translation, pelvis, L/R hips, spine1, L/R knees, spine2, L/R ankles, spine3, L/R feet, neck, L/R collars, head, L/R shoulders, L/R elbows, L/R wrists]:
\begingroup\small
\begin{equation}
\label{eq:kafs_alpha_impl}
\begin{split}
\boldsymbol{\alpha}=[&1.000,1.000,0.975,0.975,0.975,0.950,0.950,0.950,\\
&0.925,0.925,0.925,0.900,0.900,0.950,0.925,0.925,\\
&0.925,0.900,0.900,0.875,0.875,0.850,0.850].
\end{split}
\end{equation}
\endgroup
KAFS applies $\sigma_j^{(k)}=(\sigma^{(k)})^{1/\alpha_j}$ both to the per-token timestep label and to the per-token Euler increment $\sigma_j^{(k+1)}-\sigma_j^{(k)}$. In the KAFS ablations of the main paper, the uniform variant sets all $\alpha_j{=}1$ and the random variant samples $23$ values from $[0.85,1.15]$ with seed $42$.

\subsection{Compute Budget}
Using the training configuration reported in the main paper, the Motion VAE has global batch size $1024$, corresponding to $409.6$M clip presentations and $6.4$M V100 GPU-step equivalents. KU-FlowT has global batch size $480$, corresponding to $144.0$M clip presentations and $24.0$M V100 GPU-step equivalents. The combined training budget is therefore $30.4$M V100 GPU-step equivalents. At inference time, each generated segment uses $50$ KU-FlowT evaluations and one VAE decode; prefix-conditioned and segment-wise generation additionally encode the clean prefix once per conditioned segment. Because KAFS only changes the per-token noise schedule, it has the same parameter count and network-evaluation count as the base sampler.

\subsection{Evaluation preprocessing}
Our generation tables use a representation-consistent TMR evaluator. TMR~\cite{petrovich2023tmr} treats text--motion retrieval as a standalone cross-modal benchmark rather than only a proxy score, and trains a contrastive motion--text embedding space for R-Precision and matching-distance style measurements. HumanTOMATO~\cite{humantomato} further shows that text--motion alignment can benefit from a task-aligned motion--text encoder when the evaluation scope moves beyond the original body-only setting. MotionStreamer~\cite{motionstreamer} follows the same direction for SMPL animation: it evaluates motions in a 272-D SMPL-based representation and trains a TMR evaluator on processed HumanML3D-272 data. We directly use the trained TMR evaluator from MotionStreamer. We therefore keep the HumanML3D split and metric family, but score all methods after conversion to the same SMPL/MotionStreamer-272 representation so that text alignment, distributional quality, and physical statistics are computed on the motion form that is actually rendered and consumed downstream.

HumanML3D-263 baselines are evaluated in the temporal convention used by their released models before being converted to the shared TMR-evaluation protocol. In particular, methods trained on the 20\,fps HumanML3D feature representation are decoded at 20\,fps and then resampled to the 30\,fps SMPL evaluation rate. Paired reconstruction uses the same first-frame root alignment and shared forward-kinematics skeleton described in the main paper; generation tables use the realized frame lengths produced under each method's valid temporal stride.

Physical columns follow the non-VLM motion-quality metrics of MBench~\cite{vimogen} on shared SMPL-22 FK joints (metres, Y-up, horizontal axes $(x,z)$). After lowest-foot floor alignment, foot contact follows the MBench rule: 3D frame displacement below $0.01$ or height below $0.02$\,m. \emph{Slide} is mean horizontal foot displacement on contact frames, scaled by $1000$ as millimetres per frame. \emph{Float} is the percentage of frames flagged by the MBench floating heuristic over root-relative foot motion and no-contact intervals. \emph{Jitter} is global plus root-relative mean second-order joint acceleration, scaled by $1000$. Lower is better for all three columns.

\section{User Study Details}
\label{sec:user_study}

We provide the full protocol for the narrative motion composition user study reported in the main paper.

\subsection{Task description}
Narrative motion composition takes a free-form narrative description as input---containing multiple implicit actions spanning 10--30+ seconds of motion---and generates a coherent, multi-segment 3D human motion sequence. The user study compares \name with MotionStreamer~\cite{motionstreamer}, a streaming-generation baseline evaluated with the same segment plan, and KIMODO-SMPLX~\cite{kimodo}, a pose-generation baseline with a native multi-prompt interface. All compared methods first decompose the narrative into atomic action prompts and rewrite each prompt with predicted duration using a motion-aware text normalization module~\cite{hymotion}; the decomposition is frozen before any motion is generated. The same captions, durations, and realized frame counts are used for all methods, so the comparison isolates motion generation under a shared planning stage rather than differences in narrative parsing.

\subsection{Segment-plan generation}
The narrative paragraph is not passed directly to either generator. We first create a frozen segment plan shared by all evaluated methods. The text-normalization module parses explicit temporal connectives (e.g., ``then'', ``after'', ``before'', and comma-separated action chains) into an ordered list of atomic clauses. Each clause is required to contain one primary body action; simultaneous modifiers such as direction, speed, limb emphasis, and emotional style are kept inside the same clause rather than split into separate segments. The module then rewrites each clause as a standalone HumanML3D-style caption and predicts its duration in seconds. We convert duration to frame count at 30\,fps and round to each generator's valid temporal stride when needed. For each prompt, the frozen record stores the original narrative, category, ordered segment index, normalized caption, duration in seconds, realized frame count, and segment boundary metadata before any motion is generated. The ordered captions, realized frame counts, and original narrative are stored once and reused for \name, MotionStreamer, and KIMODO, so no method receives a method-specific decomposition or duration correction after generation.

\subsection{Autoregressive generation protocol}
For \name, the first segment is generated from the encoded condition of the first normalized caption using the default inference settings reported in the main paper. For segment $n{>}1$, we encode the previous segment's last $F{=}5$ frames and inject them as clean prefix tokens while denoising the target frames under the encoded text condition $c_n$. The generated prefix frames are treated as conditioning context and are dropped before concatenating the final sequence, preventing duplicated boundary frames. MotionStreamer is run with the same normalized captions and segment lengths in its streaming-continuation setting. KIMODO-SMPLX is run through its multi-prompt interface using the same segment captions and target lengths; since it does not use PRISM's clean-context/noisy-target latent mask, it serves as a controlled pose-generation baseline under the same external plan. All videos are rendered from SMPL or SMPL-X motions using the same mesh renderer, camera, frame rate, and video resolution; the videos contain no text overlays that could reveal the method identity.

\subsection{Benchmark construction}
We construct a benchmark of 50 diverse narrative prompts spanning 10 categories (5 prompts per category):
\begin{itemize}[leftmargin=5mm, itemsep=0.2ex]
\item \textbf{Daily routine}---\textit{e.g., ``A person wakes up from bed, stretches both arms overhead, yawns, walks to the bathroom, bends down to wash their face, then stands up and dries their face with a towel.''}
\item \textbf{Combat / martial arts}---\textit{e.g., ``A martial artist takes a fighting stance, throws a left jab, follows with a right cross, ducks under a swing, delivers an uppercut, then steps back into guard position.''}
\item \textbf{Dance choreography}---\textit{e.g., ``A dancer starts with arms at their sides, raises the right arm in an arc, spins on one foot, drops into a lunge, rises back up, sways hips left and right, then finishes with a dramatic pose.''}
\item \textbf{Sports / athletics}---\textit{e.g., ``A gymnast runs forward, hurdles into a round-off, immediately does a back handspring, lands and takes a step, raises both arms in a salute, then walks back.''}
\item \textbf{Performance / theater}---\textit{e.g., ``A mime pushes against an invisible wall on the left, then the right, then overhead, crouches down, discovers an invisible door handle, turns it, and steps through.''}
\item \textbf{Exercise / fitness}---\textit{e.g., ``A person does five jumping jacks, drops for three push-ups, jumps to standing, performs two squat jumps, then rests with hands on knees.''}
\item \textbf{Social interaction}---\textit{e.g., ``A person walks up, waves, joins a conversation circle, nods along, gestures broadly, laughs and leans back, looks at a watch, and waves goodbye.''}
\item \textbf{Occupational}---\textit{e.g., ``A waiter carries an invisible tray, stops, carefully lowers it, sets items on the table, straightens up, tucks the tray under the arm, and walks away.''}
\item \textbf{Adventure / exploration}---\textit{e.g., ``A hiker walks uphill, pulls themselves up a steep section, reaches the top, shields eyes to look into the distance, takes a deep breath with arms wide, then begins walking along the ridge.''}
\item \textbf{Narrative story}---\textit{e.g., ``A person hears a noise, freezes, slowly turns to look behind, sees something alarming, takes two quick steps back, turns and sprints forward, then dives to the ground and covers their head.''}
\end{itemize}
Each prompt describes 4--10 sub-actions. The benchmark is designed to stress-test autoregressive chaining, segment stitching, and the full decomposition pipeline.

\subsection{Evaluation procedure}
For each of the 50 prompts, each evaluated method generates a motion video rendered from the same viewpoint with identical mesh visualization. The completed quantitative study presents anonymous, randomized triplets containing \name, MotionStreamer, and KIMODO-SMPLX. Evaluators rank the three videos for each criterion. The ranking interface allows ties, so a valid annotation may be a strict order, such as \name $>$ KIMODO $>$ MotionStreamer, or a tied order, such as \name $=$ KIMODO $>$ MotionStreamer.

\subsection{Evaluation dimensions and protocol}
For each triplet, evaluators rank the three methods on four dimensions:
\begin{itemize}[leftmargin=5mm, itemsep=0.2ex]
\item \textbf{Motion Quality}: naturalness and physical plausibility of the generated motion (absence of jitter, foot sliding, ground penetration, and unnatural poses).
\item \textbf{Text Fidelity}: semantic alignment between the narrative description and the generated motion (whether all described actions are correctly performed in the right order).
\item \textbf{Transition Smoothness}: coherence at segment boundaries (whether transitions between consecutive actions appear smooth and natural, without abrupt jumps or freezing).
\item \textbf{Overall Preference}: overall quality judgment considering all of the above.
\end{itemize}
The rank protocol reduces subjective noise compared to Likert-scale scoring: evaluators make relative comparisons within the same prompt, while ties avoid forcing distinctions when two videos are not reliably separable.

\subsection{Participants}
20 evaluators participate in the study. All are graduate students or researchers with experience in computer vision or computer graphics. Each evaluator rates all 50 triplets, yielding 1{,}000 rank annotations per dimension (20 evaluators $\times$ 50 prompts). The study takes approximately 30--40 minutes per evaluator.

\subsection{Statistical analysis}
We convert each three-method rank annotation into PRISM-centered pairwise outcomes. The user-study table in the main paper reports the two pairs, \name--KIMODO and \name--MotionStreamer. For each listed pair, Win means that \name is ranked above the other method, Tie means that they share the same rank, and Loss means that \name is ranked below the other method. For each pair and criterion, we normalize the pairwise outcomes over all 1{,}000 annotations, so the reported W/T/L percentages sum to 100.0\%. We apply a two-sided sign test over non-tie pairwise outcomes; differences described as significant pass $p<0.01$, while pairs with similar Win and Loss rates are treated as comparable under the tie-aware rank protocol.

\end{document}